\documentclass{bmvc2k}

\title{Multispectral Deep Neural Networks for Pedestrian Detection}

\addauthor{Jingjing Liu}{http://paul.rutgers.edu/~jl1322}{1}
\addauthor{Shaoting Zhang}{szhang16@uncc.edu}{2}
\addauthor{Shu Wang}{sw498@cs.rutgers.edu}{1}
\addauthor{Dimitris N. Metaxas}{https://www.cs.rutgers.edu/~dnm}{1}

\addinstitution{
 Department of Computer Science\\
 Rutgers University\\
 Piscataway, NJ, USA
}
\addinstitution{
 Department of Computer Science\\
 UNC Charlotte\\
 Charlotte, NC, USA
}

\runninghead{Liu et al.}{Multispectral DNNs for pedestrian detection}

\def\eg{\emph{e.g}\bmvaOneDot}

\def\etal{\emph{et al}\bmvaOneDot}
\def\vs{\emph{vs}\bmvaOneDot}
\newcommand{\myparagraph}[1]{\smallskip\noindent\textbf{#1}} 
\usepackage{wrapfig}
\usepackage{color}
\usepackage[outdir=./images/]{epstopdf}

\begin{document}

\maketitle

\begin{abstract}
Multispectral pedestrian detection is essential for around-the-clock applications, \eg, surveillance and autonomous driving. We deeply analyze Faster R-CNN for multispectral pedestrian detection task and then model it into a convolutional network (ConvNet) fusion problem. Further, we discover that ConvNet-based pedestrian detectors trained by color or thermal images separately provide complementary information in discriminating human instances. Thus there is a large potential to improve pedestrian detection by using color and thermal images in DNNs simultaneously. We carefully design four ConvNet fusion architectures that integrate two-branch ConvNets on different DNNs stages, all of which yield better performance compared with the baseline detector. Our experimental results on KAIST pedestrian benchmark show that the Halfway Fusion model that performs fusion on the middle-level convolutional features outperforms the baseline method by 11\% and yields a missing rate 3.5\% lower than the other proposed architectures.
\end{abstract}

\section{Introduction}
\label{sec:intro}
As a canonical case of general object detection problem, in the past decades, pedestrian detection has attracted consistent attention from computer vision community~\cite{dalal2005hog, wang2009hog, benenson2012ped, felzenszwalb2010, mathias2013, marin2013, wang2014st, nam2014local, cai2015learning, paisitk2014, liu2016people}. It is the principle technique for various applications, such as surveillance, tracking, autonomous driving, etc. Although numerous efforts have been made for this problem and significant improvement has been achieved in recent years~\cite{benenson2014ten}, there still exists an insurmountable gap between current machine intelligence and human perception ability on pedestrian detection~\cite{zhang2016far}. Many challenges prevent artificial vision systems from practical applications, including occlusion, low image resolution, and cluttered background. Besides, since most of current pedestrian detectors explored color images of good lighting, they are very likely to be stuck with images captured at night, due to bad visibility of pedestrians. Such defect would cut these approaches off from the around-the-clock applications, \eg, self-driven car and surveillance system.

\begin{figure}
\centering
\includegraphics[width=0.95\textwidth, height=0.35\textwidth]{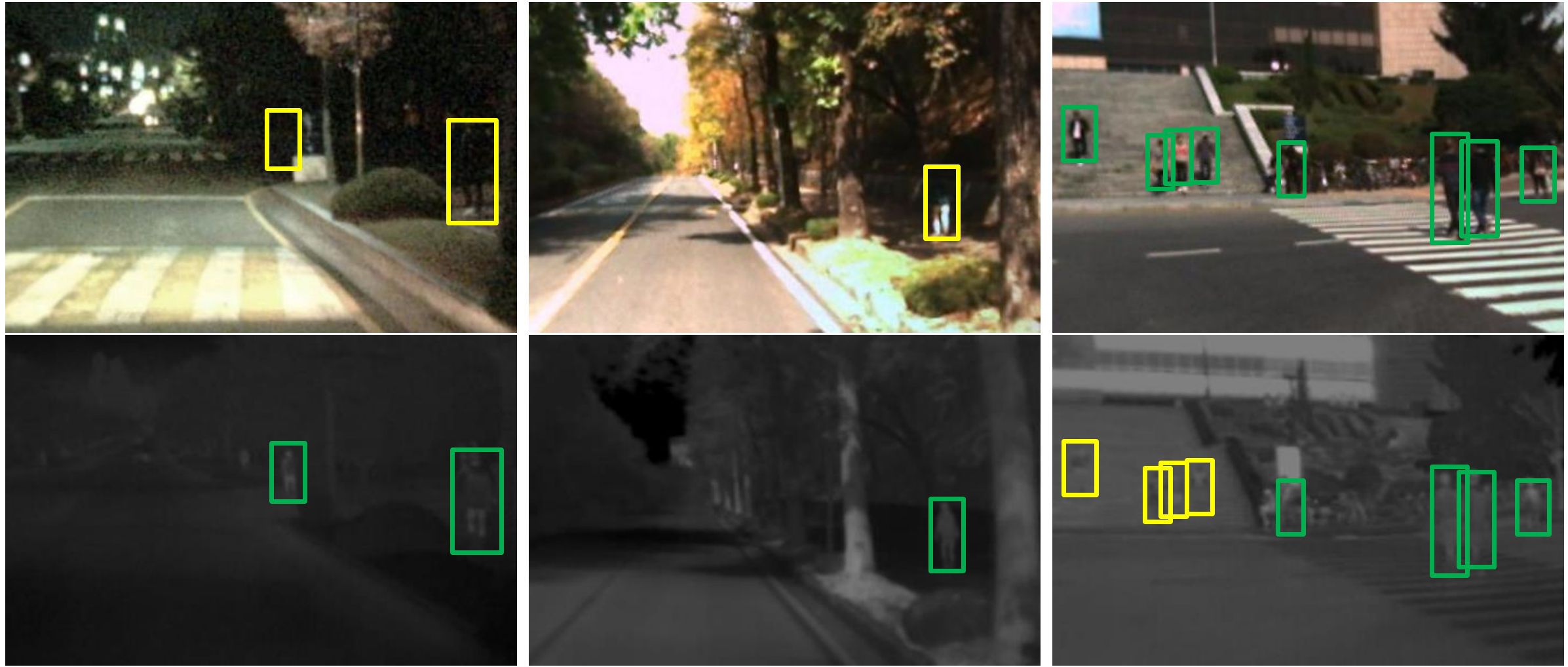}
\caption{Left and middle: the thermal images capture human shapes even in bad lighting, while the corresponding color images are messed up. Right: with bright background, color image provides more distinctive visual features for the pedestrians (standing on stairs) against background objects; in such scenario, human silhouettes in the thermal image are ambiguous. Yellow bounding boxes indicate detection failures with one image channel.} 
\vspace{-10pt}
\label{fig:1}
\end{figure}

In fact, aforementioned difficulties not only exist in pedestrian detection, but also in many other vision tasks. To tackle these problems, other types of sensors beyond RGB cameras were developed, such as depth cameras (time-of-flight, or near infrared, \eg, Kinect) and thermal cameras (long-wavelength infrared). Since ambient lighting has less effect on thermal imaging, thermal cameras are widely used in human tracking~\cite{torabi2012}, face recognition~\cite{sarfraz2015deep}, and activity recognition~\cite{han2005human} for its robustness. With regard to pedestrian detection, thermal images usually present clear silhouettes of human objects~\cite{han2007fusion,socarras2011}, but lose fine visual details of human objects (\eg, clothing) which can be captured by RGB cameras (depending on external illumination). As shown in Figure~\ref{fig:1}, instances of yellow bounding boxes would fail in detection with one image channel (color or thermal), while the other might help. 

In some sense, color and thermal image channels provide complementary visual information. 
Nevertheless, except very recent effort~\cite{gonzalez2016, wagner2016}, most of previous studies focus on detecting pedestrians with color or thermal images only, rather than leveraging color and thermal images simultaneously. 
Furthermore, although researchers have applied deep neural networks (DNNs) on vision problems with multimodal data sources, \eg, action recognition~\cite{simonyan2014}, image classification~\cite{srivas2012multimodal}, etc., it is still unknown how color and thermal image channels can be properly fused in DNNs to achieve the best pedestrian detection synergy. 

In this paper, we focus on how to make the most of multispectral images (color and thermal) for pedestrian detection. 
With the recent success of DNNs~\cite{Goodfellow2016} on generic object detection~\cite{girshick2014}, it becomes very natural and interesting to exploit the effectiveness of DNNs for multispectral pedestrian detection. 
We first adapt Faster R-CNN~\cite{ren2015faster} model into our vanilla convolutional network (ConvNet), and train two separate pedestrian detectors on color or thermal images, respectively. 
It is not surprising to discover that these two ConvNet-based detectors provide complementary detection decisions, and there is a large potential to improve the detection performance by leveraging multispectral images, especially for around-the-clock applications. 
However, it is not trivial to explore the most effective DNNs architecture that simultaneously utilizes color and thermal images for pedestrian detection.
Then the challenge of multispectral pedestrian detection task becomes a ConvNet fusion problem. 
From the perspective of fusions on different DNNs levels, four ConvNet fusion architectures are carefully designed upon our vanilla ConvNet, and then well investigated with extensive experimental evaluation.
Our major contribution is fourfold:
\begin{itemize}
\vspace{-4pt}
\item First, we carefully design four distinct ConvNet fusion architectures that integrate two-branch ConvNets on different DNNs stages, \emph{i.e.}, convolutional stages, fully-connected stages, and decision stage, corresponding to information fusion on low level, middle level, high level, and confidence level. All these models outperform the strong baseline detector Faster R-CNN on KAIST multispectral pedestrian dataset (KAIST)~\cite{hwang2015mul}.
\vspace{-4pt}
\item Second, we reveal that our Halfway Fusion model -- fusion of middle-level convolutional features, provides the best performance on multispectral pedestrian detection. This implies that the best choice of fusion scheme is a balance between fine visual details and semantic information.
\vspace{-4pt} 
\item Third, our Halfway Fusion model significantly reduces the missing rate of baseline method Faster R-CNN by 11\%, yielding a 37\% overall missing rate on KAIST, which is also 3.5\% lower than the other proposed fusion models. 
\vspace{-4pt}
\item Last but not least, our vanilla ConvNet achieves state-of-art performance (17\% missing rate) on Caltech pedestrian benchmark~\cite{dollar2009ped}. As far as we know, this is the first time Faster R-CNN~\cite{ren2015faster} has been investigated for pedestrian detection.
\end{itemize}

\section{Related Work}
\label{sec:rel}
\myparagraph{DNNs for Pedestrian Detection:} One pioneer work used deep neural network for pedestrian detection was proposed in~\cite{sermanet2013ped}, which combined multi-stage unsupervised feature learning with multi-scale DNNs. In~\cite{ouyang2013}, Ouyang \etal modeled visibility relationships among pedestrians using DNNs, which improved the visual confidences of human parts.
Tian \etal~\cite{tian2015deep} trained 45 complementary part detectors with weakly annotated humans, to handle partial occlusions. Tian \etal~\cite{tian2015ped} improved pedestrian detection by learning high-level features from DNNs of multiple tasks, including pedestrian attribute prediction. Angelova \etal~\cite{angelov2015} built DNNs cascades that filtered candidates by tiny DNNs and steed up detection to 15 FPS. Li \etal~\cite{li2015scale} proposed scale-aware DNNs with a scale gate function, to capture characteristic features for pedestrians of different image sizes. Recent DNNs-based pipelines focused on incorporating DNNs with some pedestrian detectors~\cite{dollar2012ped, benenson2014ten} that were used to generate class-specific proposals. These proposals were then passed into DNNs for classification. Along this research stream, many DNNs architectures have been investigated, including CifarNet~\cite{hosang2015}, AlexNet~\cite{krizhevsky2012}, and VGG-16~\cite{simonyan2014very}. General speaking, deeper DNNs performed better than shallow DNNs. However, these methods depended largely on qualities of proposals from a pre-trained pedestrian detector. In other words, a good detector should produce qualified hard negatives for training and enough positives for testing as many as possible.

\myparagraph{DNNs with Multimodal inputs:} It is an essential challenge for many vision problems to integrate information from multimodal data sources. A bunch of DNNs-based multimodal models have been proposed, involving in data sources of different modalities, such as image \vs audio~\cite{ngiam2011multimodal}, image \vs text~\cite{srivas2012multimodal, wang2015learning}, image \vs video~\cite{simonyan2014, karpathy2014}, etc. In ~\cite{ngiam2011multimodal}, Naiam \etal learned a hidden layer from deep belief networks (DBNs) as the shared representations of videos and audios. Similar network was also applied in~\cite{srivas2012multimodal} for image classification and retrieval, while missing modality was tolerated. Wang \etal~\cite{wang2015learning} imposed structure-preserving constraints into their similarity objective function for images and texts, achieving better phrase localization results in images. For video recognition, Simonyan \etal~\cite{simonyan2014} proposed two-stream Covnets that incorporated spatial and temporal information. In their method, optical flow ConvNet was combined with key frame based ConvNets. Color and depth images have also been exploited simultaneously for 3D object classification~\cite{socher2012cnn} and 3D object detection~\cite{gupta2014}. Most of the aforementioned methods used two-branch network and then fused features from different channels at very end, \emph{i.e.}, last feature layer fusion~\cite{socher2012cnn, gupta2014, wang2015learning, yan2015deep} or confidence fusion~\cite{simonyan2014}.  Karpathy \etal~\cite{karpathy2014} discussed some fusion schemes for video classification. However, DNNs-based multispectral pedestrian detection has not been studied thoroughly. It is still an open question that how color and thermal image channels could be fused properly in DNNs for pedestrian detection, to obtain `optimal' synergy.

\section{Methodology}
\label{sec:method}

\subsection{Vanilla ConvNet}
\label{sec:vanilla}
There exists many ConvNet architectures that are applicable for multispectral pedestrian detection. Some of them have been discussed in Section~\ref{sec:rel}. Inspired by the recent success on general object detection, we consider starting with Faster R-CNN~\cite{ren2015faster} and verify its performance on Caltech pedestrian benchmark~\cite{dollar2009ped}. 

Faster R-CNN model is consisted of a Region Proposal Network (RPN) and a Fast R-CNN detection network~\cite{girshick2015fast}. RPN is a fully convolutional network that shares convolutional features with the detection network. Compared to its methodological ancestries, \emph{i.e.}, R-CNN~\cite{girshick2014} and Fast R-CNN~\cite{girshick2015fast}, Faster R-CNN could produce proposals of high-quality by RPN. This is different from other ConvNet-based pedestrian detectors~\cite{hosang2015, li2015scale} that reply an independent proposal generators. RPN enables nearly cost-free region proposals which generates 300 proposals in less than 0.3s. Besides, Faster R-CNN uses Region of Interest (RoI) Pooling layer that pools the feature map of each proposal into a fixed size, \emph{i.e.},  $7\times 7$, thus could handle pedestrians of arbitrary sizes. 

\begin{wrapfigure}{r}{0.5\textwidth}
\vspace{-8pt}
  \centering
 \includegraphics[width=0.48\textwidth]{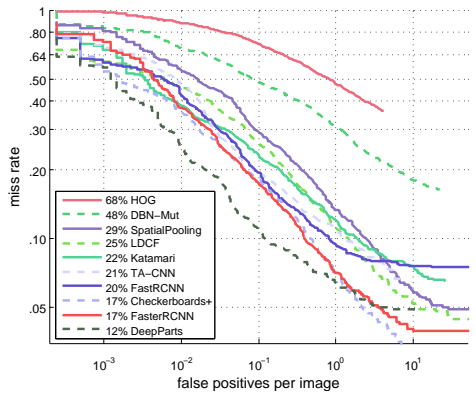}
  \vspace{-5pt}
  \caption{Comparison of detection results reported on the test set of Caltech pedestrian benchmark. Our vanilla ConvNet achieved $17\%$ MR. }
  \label{fig:caltech}
\end{wrapfigure}

\myparagraph{Implementation details:} We exploited the original Faster R-CNN model with a few twists and adapt it into our vanilla ConvNet. We removed the fourth max pooling layer of the very deep VGG-16 model~\cite{simonyan2014very}. This is encouraged by the observation in~\cite{li2015scale} that larger feature maps are beneficial for detecting pedestrians of small image sizes. Faster R-CNN uses reference anchors of multi-scale ($\times 3$) and multi-ratios ($\times 3$) to predict locations of region proposals. Given the typical aspect ratio of pedestrians, we discarded the anchor ratio 0.5 to accelerate the training and testing of RPN. 

Caltech$\times$10 training set was used for fine-tuning. We excluded occluded, truncated, and small ($<$ 50 pixels) pedestrian instances, resulting in around 7000 training images. Proposals of intersection-over-union (IoU) with any ground truth pedestrians larger than 0.5 were regarded as positives, otherwise they were used as negative samples. Following the alternative training routine of Faster R-CNN, the networks were initialized by the pre-trained VGG-16 model and then fine-tuned with Stochastic Gradient Descent (SGD) for about 6 epochs. The learning rate (LR) was set to 0.001 and reduced to 0.0001 after 4 epochs. Single image scale (600 pixels) was used, without considering feature pyramids.

\myparagraph{Comparison of Detections:} We compared our vanilla ConvNet (FasterRCNN) with some other methods reported on Caltech test set, including HOG~\cite{dalal2005hog}, DBN-Mut~\cite{ouyang2013}, SpatioPooling~\cite{paisitk2014}, LDCF~\cite{ nam2014local}, Katamari~\cite{benenson2014ten}, TA-CNN~\cite{tian2015ped}, FastRCNN~\cite{girshick2015fast}, Checkerboards+~\cite{zhang2015filtered}, and DeepParts~\cite{tian2015deep}. For FastRCNN, we used ACF pedestrian detector~\cite{dollar2014fast} to obtain proposals which were then used to train and test Fast R-CNN detection network. A low threshold (-50) was set for ACF detector, so as to produce enough proposals. We used IoU 0.5 to validate detections. Detection performance was measured by log-average miss-rate over the range of $[10^{-1},10^0]$ (MR, lower is better) under reasonable configuration~\cite{dollar2012ped}. As shown in Figure ~\ref{fig:caltech}, with completely data-driven and end-to-end training, our vanilla ConvNet beat most of state-of-art approaches that depend on sophisticated features or network design. 
We achieved 17\% MR, which is lower than some DNNs-based methods, such as DBN-Mut(48\%), TA-CNN(21\%), and FastRCNN (20\%). With a single ConvNet, the performance of our vanilla ConvNet is approaching DeepParts (12\%) that is an assembly of 45 part ConvNets.

Given its state-of-art performance and some merits, \eg, end-to-end training and capability of handling pedestrians of arbitrary sizes, in our paper, the vanilla ConvNet is used as the fundamental DNNs architecture in designing multispectral detectors.

\subsection{Multispectral ConvNets}
Intuitively, color and thermal image channels provide auxiliary visual information to each other in depicting pedestrian objects. If one image channel fails in detection, \emph{i.e.}, missing true detections or recalling false alarms, the other could still make the correct decision. 
In this section, first of all, we are trying to answer the following questions: when strong ConvNet-based detectors are involved, does color and thermal images still provide complementary information? To what extend the improvement should be expected by fusing them together? Next, we model the multispectral pedestrian detection task as a ConvNet fusion problem. We carefully design four distinct ConvNet fusion models that integrate two-branch ConvNets at different DNNs stages. Each model represent one multispectral pedestrian detector.

\subsubsection{Are They Really Complementary?}
To study the complementary potential between color and thermal images, we first trained two separate pedestrian detectors with color or thermal images only, based on our vanilla ConvNet, namely FasterRCNN-C and FasterRCNN-T. The training set of KAIST dataset~\cite{hwang2015mul} (more details will be given in Section~\ref{sec:exp}) was used in fine-tuning of neural networks, while the results were validated on the test images. In our following analyses, only detections of more than 0.5 confidence scores were considered. We regarded detections of IoUs with any ground truth (GT) larger than 0.5 as true positives (TPs), otherwise as false positives (FPs). Multiple detections on the same GT were treated as FPs. In Table~\ref{tab:1}, we enumerate the numbers of GTs, TPs, and FPs of FasterRCNN-C and FasterRCNN-T, in terms of all-day, daytime, and nighttime images. TP$_{(C,T)}$ denotes pedestrians detected by both of the two detectors. TP$_{(C)}$ and TP$_{(T)}$ represent instances exclusively detected by FasterRCNN-C or FasterRCNN-T. Analogously, we have FP$_{(C,T)}$, FP$_{(C)}$, and FP$_{(T)}$ for false alarms.
\begin{table}[h]
\centering
\begin{tabular}{|c||c|c|c|c||c|c|c|}
\hline
{} & {GT} & {TP$_{(C,T)}$} & {TP$_{(C)}$} & {TP$_{(T)}$} & 
{FP$_{(C,T)}$} & {FP$_{(C)}$} & {FP$_{(T)}$}\\ \hline
{All} & {2,757} & {924} & {390} & {397} & {345} & {1,169} & {1,158}\\ \hline
{Day} & {2,003} & {720} & {346} & {176} & {303} & {745} & {827}\\ \hline
{Night} & {754} & {204} & {44} & {221} & {42} & {424} & {331}\\ \hline
\hline
\end{tabular}
\vspace{2pt}
\caption{Numbers of ground truths, true detections, and false alarms reported on the test images of KAIST pedestrian dataset, in terms of all-day, daytime, and nighttime, respectively.}
\label{tab:1}
\end{table}
\vspace{-2pt}

Obviously, FasterRCNN-C and FasterRCNN-T have consensuses on pedestrian detections to some extent, but not alway. They have overall 924 common TPs, while 390 pedestrians captured by FasterRCNN-C were regarded as background by FasterRCNN-T. In daytime, FasterRCNN-C gets more TPs than FasterRCNN-T (1,066 \vs 896), while the trend is opposite on nighttime images (248 \vs 425). It is reasonable that during daytime most pedestrians are in good lighting conditions, except some corner cases (standing in shadow), when thermal images are apt to be affected by sunlight. In contrast, thermal images could capture better visual features of pedestrians at night. Besides, FasterRCNN-C and FasterRCNN-T share relatively fewer FPs (345), while they get FPs 2,672 in total. It is not hard to infer that there is a large potential in excluding FPs by using two image channels.

Without doubt, color and thermal images provide complementary information on pedestrian detection. Based on the 2,252 test images, if we make an extreme assumption that all true detections from either FasterRCNN-C or FasterRCNN-T were kept and only shared false alarms were retained, the detection rate can be increased from 47.9\% to 62.1\%, with the FPPI (false positives per image) reduced from 0.549 to 0.125. Hence, we should pay serious attention on the potential improvement that would be raised from multispectral detection.
\begin{figure}[b]
\centering
\includegraphics[width=0.9\textwidth]{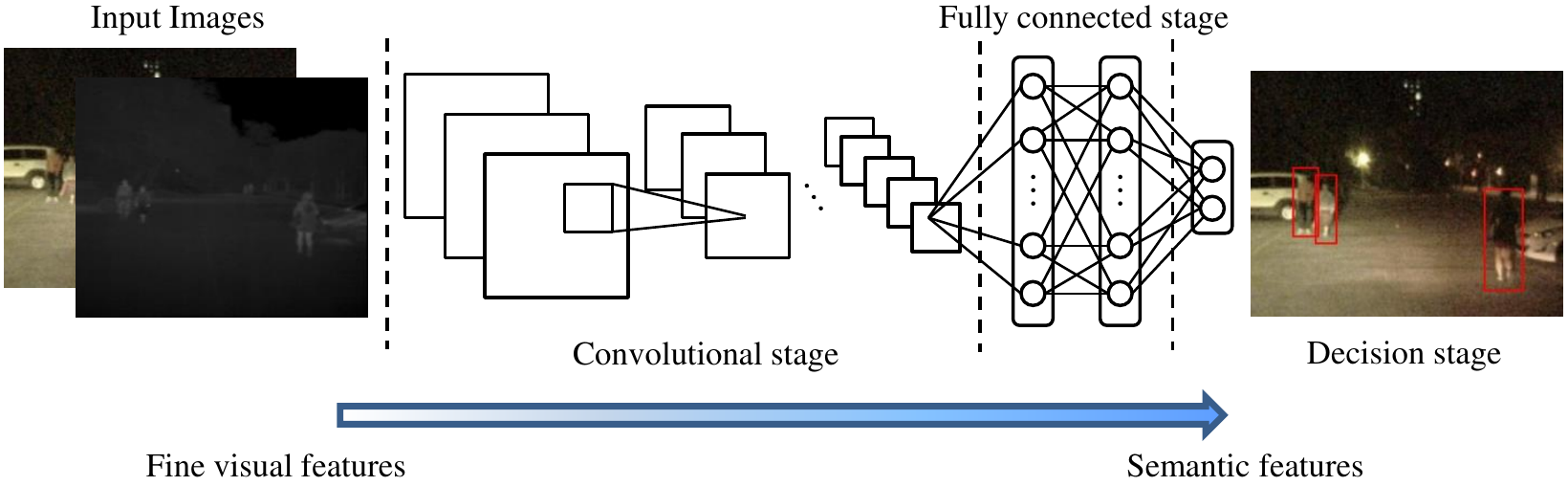}
\caption{Different stages in a ConNet. Features at different stages correspond to various levels of semantic meanings and fine visual details.} 
\label{fig:stages}
\vspace{-10pt}
\end{figure}

\subsubsection{ConvNet Fusion Models}
\begin{wrapfigure}{r}{0.485\textwidth}
\centering
\vspace{-8pt}
\includegraphics[width=0.45\textwidth]{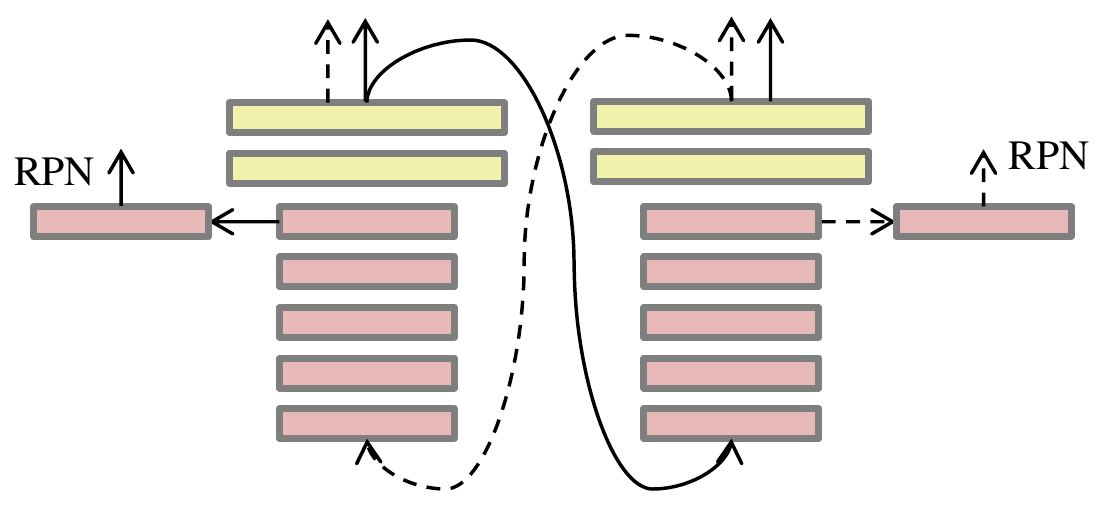}
\caption{Explored approach for combining scores of two ConvNets (Score Fusion).} 
\label{fig:score}
\vspace{-8pt}
\end{wrapfigure}

\begin{figure}[t]
\centering
\includegraphics[width=0.99\textwidth]{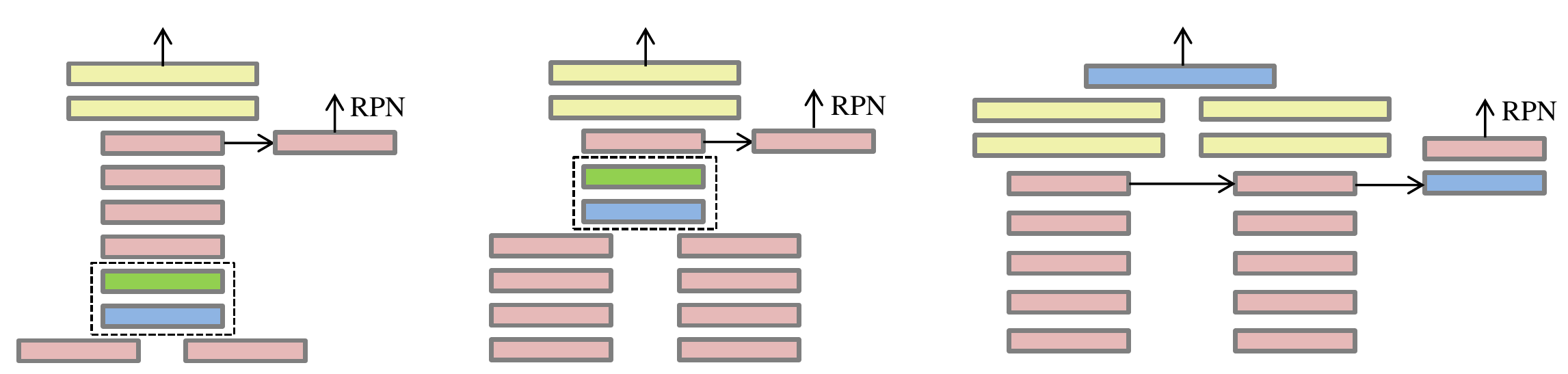}
\caption{Explored approaches to fuse color and thermal images for multispectral pedestrian detection. These approaches implement feature fusions. From left to right are feature fusions at low level (Early Fusion), middle level (Halfway Fusion), and high level (Late Fusion), respectively. Red and yellow boxes represent convolutional and fully-connected layers. Blue boxes represent concatenate layer. Green boxes denote Network-in-Network (NIN) used for dimension reduction. For the sake of conciseness, ReLU layers, pooling layers, and dropout layers are hidden from view in this figure. (Best viewed in color.)} 
\label{fig:fusion}
\vspace{-8pt}
\end{figure}

The question now is how a good multispectral pedestrian detector that explores color-thermal image pairs could be achieved. As shown in Figure~\ref{fig:stages}, a ConvNet-based detector is composed of three stages: the convolutional stage, the fully-connected stage, and the decision stage. Features at different stages corresponding to various levels of semantic meanings and fine visual details. We think fusion at different stages would lead to different detection results. Therefore, the multispectral pedestrian detection task comes to be a ConvNet fusion problem, \emph{i.e.}, what architecture of the fusion model could get best detection synergy. To this end, we make thorough inquiries on four fusion models designed upon our vanilla ConvNet. Basically, they are two-branch ConvNet models that perform fusions at different stages, denoted as Early Fusion, Halfway Fusion, Late Fusion, and Score Fusion. Three of them implement feature fusions, as shown in Figure~\ref{fig:fusion}, while the other one combines confidence scores from color and thermal ConvNet branches at decision stage, as shown in Figure~\ref{fig:score}.

\myparagraph{Early Fusion} concatenates the feature maps from color and thermal branches immediately after the first convolutional layers (C1).  Afterwards, we introduce Network-in-Network (NIN)~\cite{lin2013network, szegedy2015gnet} after feature concatenation, which is actually a $1\times1$ convolutional layer. NIN reduces the dimension of concatenate layer to 128, such that other filters from the pre-trained VGG-16 model can be reused. Besides, NIN outputs linear combinations of local features from color and thermal branches. Followed by ReLU, it can enhance the discriminability of local patches. Since C1 captures low-level visual features, such as corners, line segments, etc., Early Fusion model fuses features at low-level.

\myparagraph{Halfway Fusion} also implements fusion at convolutional stage. Different from Early fusion, it puts the fusion module after the fourth convolutional layers (C4). NIN is also used after the concatenate layer, for the same reasons as discussed before. Features from C4 layers contain more semantic meanings than C1 features, while retaining some fine visual details.

\myparagraph{Late Fusion} concatenates the last fully-connected layers (F7), which performs feature fusion at fully-connected stage. Conventionally, F7 features are used as new representations of objects.  Late Fusion executes high-level feature fusion. To be noticed, RPN here exploits C5 features from the two branches to predict human proposals. 

\myparagraph{Score Fusion} can be regarded as a cascade of two ConvNets (Figure~\ref{fig:score}). We first get detections from color ConvNet which are then sent into the other ConvNet to obtain detection scores based on thermal image, and vice verse. In practice, this can be  accomplished by using RoI Pooling layer. Detection scores from the two branches are merged with equal weights (\emph{i.e.}, 0.5). 

\section{Experiments}

\label{sec:exp}
\myparagraph{Dataset:} KAIST multispectral pedestrian dataset (KAIST)~\cite{hwang2015mul} contains $95,328$ aligned color-thermal frame pairs, with $103,128$ dense annotations on $1,182$ unique pedestrians. We sampled images from training videos with 2-frame skips, and finally obtained 7,095 training images binded with qualified pedestrians (the same criteria as discussed in Section~\ref{sec:vanilla}). The test set of KAIST contains 2,252 images sampled from test videos with 30-frame skips, among which 1,455 images were captured during daytime and 797 others for nighttime.

\myparagraph{Implementation Notes:} Most of filters in the ConvNet fusion models were initialized by the corresponding parameters of the pre-trained VGG-16 model, except new introduced layers. For instance, in Early Fusion and Halfway Fusion, the weights of NINs were initialized by a Gaussian distribution. Parallel branches in the four fusion models did not share weights. The two branches in Early Fusion, Halfway Fusion, and Late Fusion were trained simultaneously, while the two ConvNets in Score Fusion were trained individually. All the models were fine-tuned with SGD for 4 epochs with LR 0.001 and 2 more epochs with LR 0.0001. Besides, non-maximum suppression (NMS) was applied to the detections of Score Fusion model, in order to avoid double detections from color and thermal channels.
\begin{figure}
\centering
\subfigure
{\includegraphics[width=0.32\textwidth]{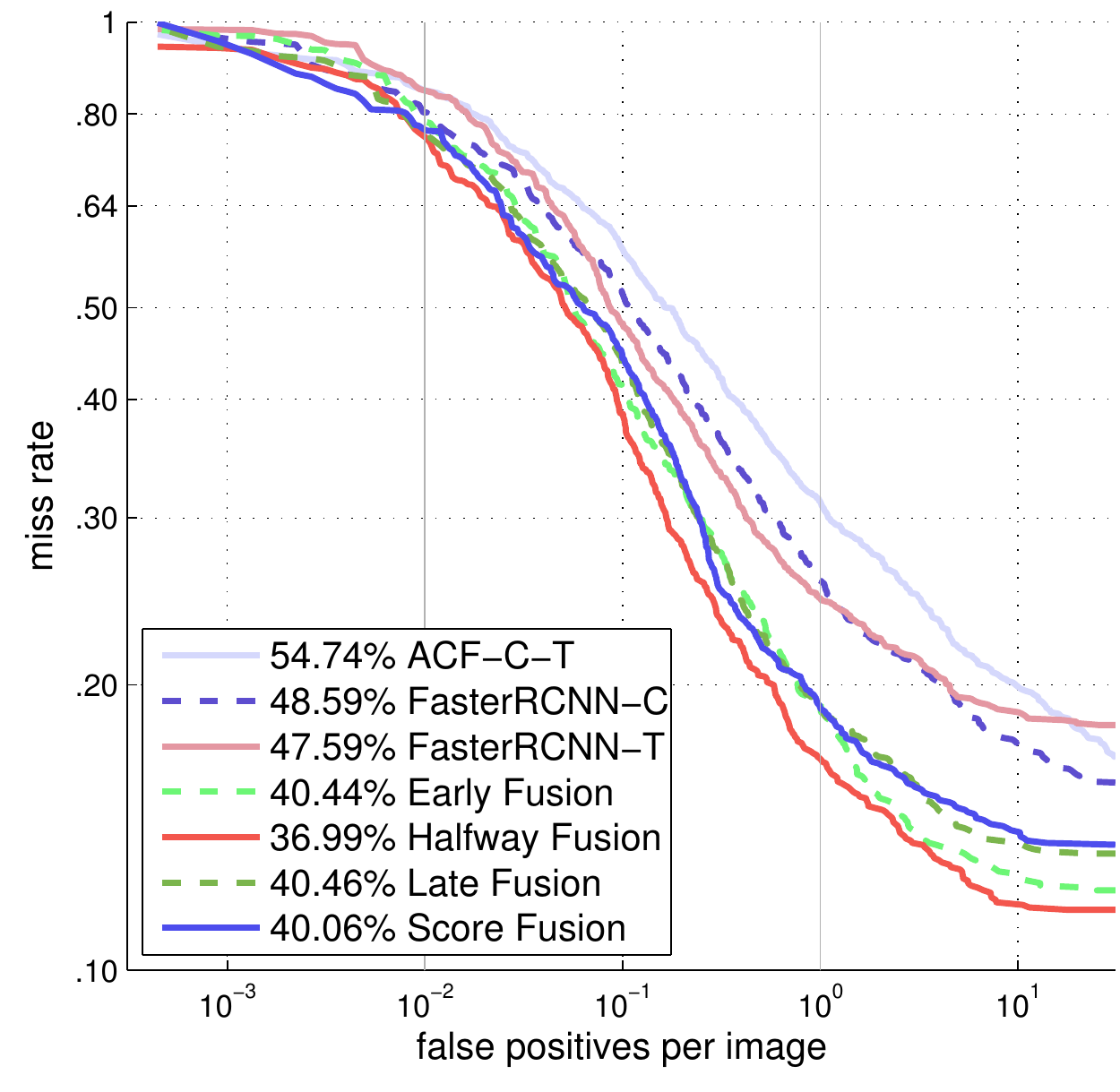}
\label{fig:deta}}
\subfigure
{\includegraphics[width=0.32\textwidth]{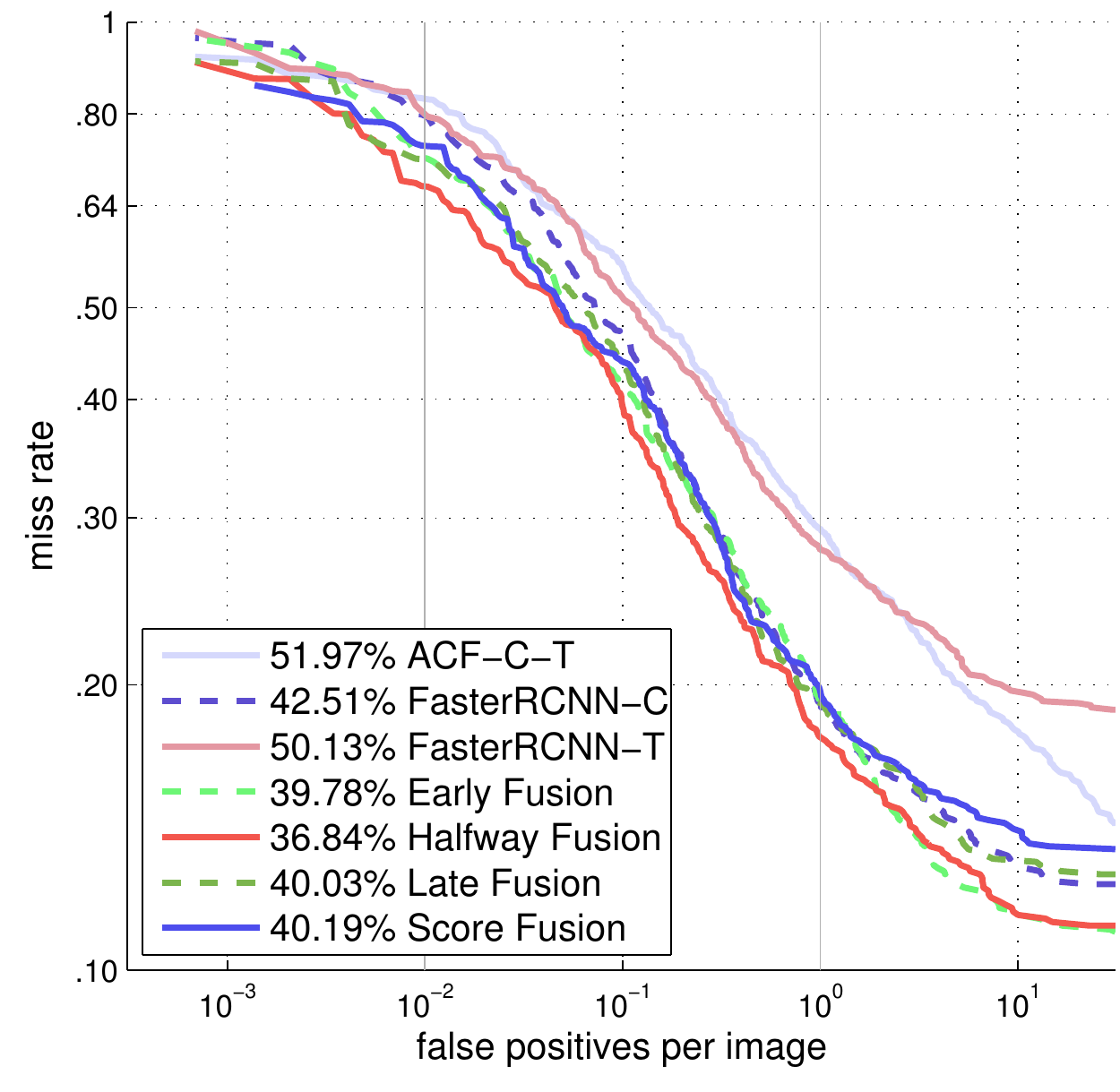}
\label{fig:detb}}
\subfigure
{\includegraphics[width=0.32\textwidth]{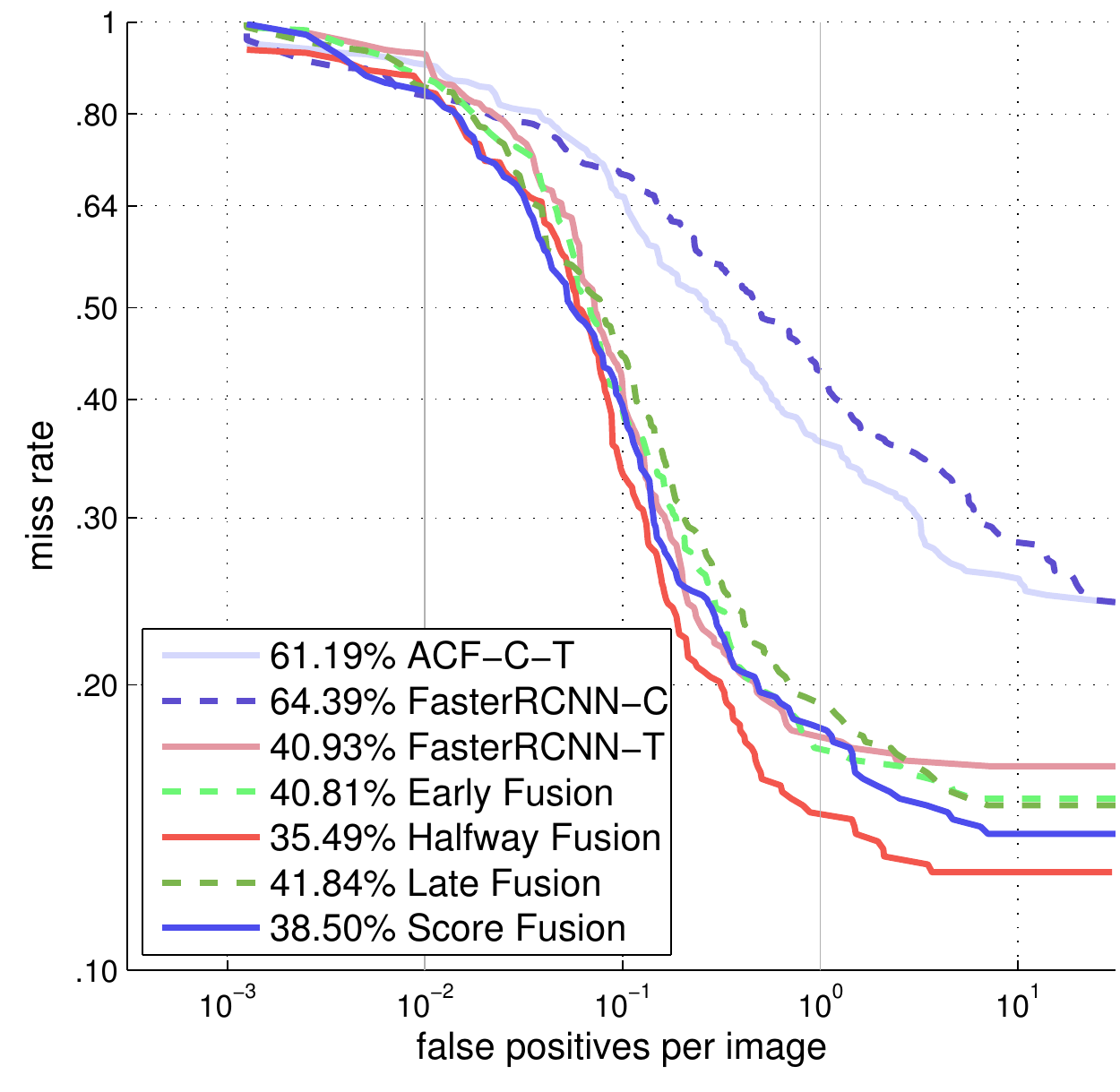}
\label{fig:detc}}
\caption{Comparison of detection results reported on the test set of KAIST multispectral pedestrian dataset, in terms of all-day (left), daytime (middle), and nighttime (right).}
\label{fig:det}
\end{figure}

\subsection{Evaluation of Detections:}
We evaluated the proposed four ConvNet fusion models on the test set of KAIST, compared to FasterRCNN-C and FasterRCNN-T, as well as ACF-C-T detector reported in~\cite{hwang2015mul}. The ACF-C-T detector used 10-channel aggregated features to fuse color and thermal images. Comparisons of detection results are presented in Figure~\ref{fig:det}, in terms of MR under reasonable configuration~\cite{dollar2012ped}.

Generally speaking, detectors with single image modality obtain inferior results than fusion models. FasterRCNN-C obtains 42.5\% MR on daytime images, while working worse than the ACF-C-T detector (64.4\% \vs 61.2\%) on nighttime images. On the other side, FasterRCNN-T suffers on daytime images, although it gets similar MR as some fusion models on nighttime images. Consequently,  both FasterRCNN-C and FasterRCNN-T are not applicable for around-the-clock applications. Compared to FasterRCNN-C and FasterRCNN-T, ConvNet fusion models produce significantly better results, which reduce the overall MR from 48\% to around 40\%.

\begin{figure}[h]
\centering
\includegraphics[width=0.98\textwidth]{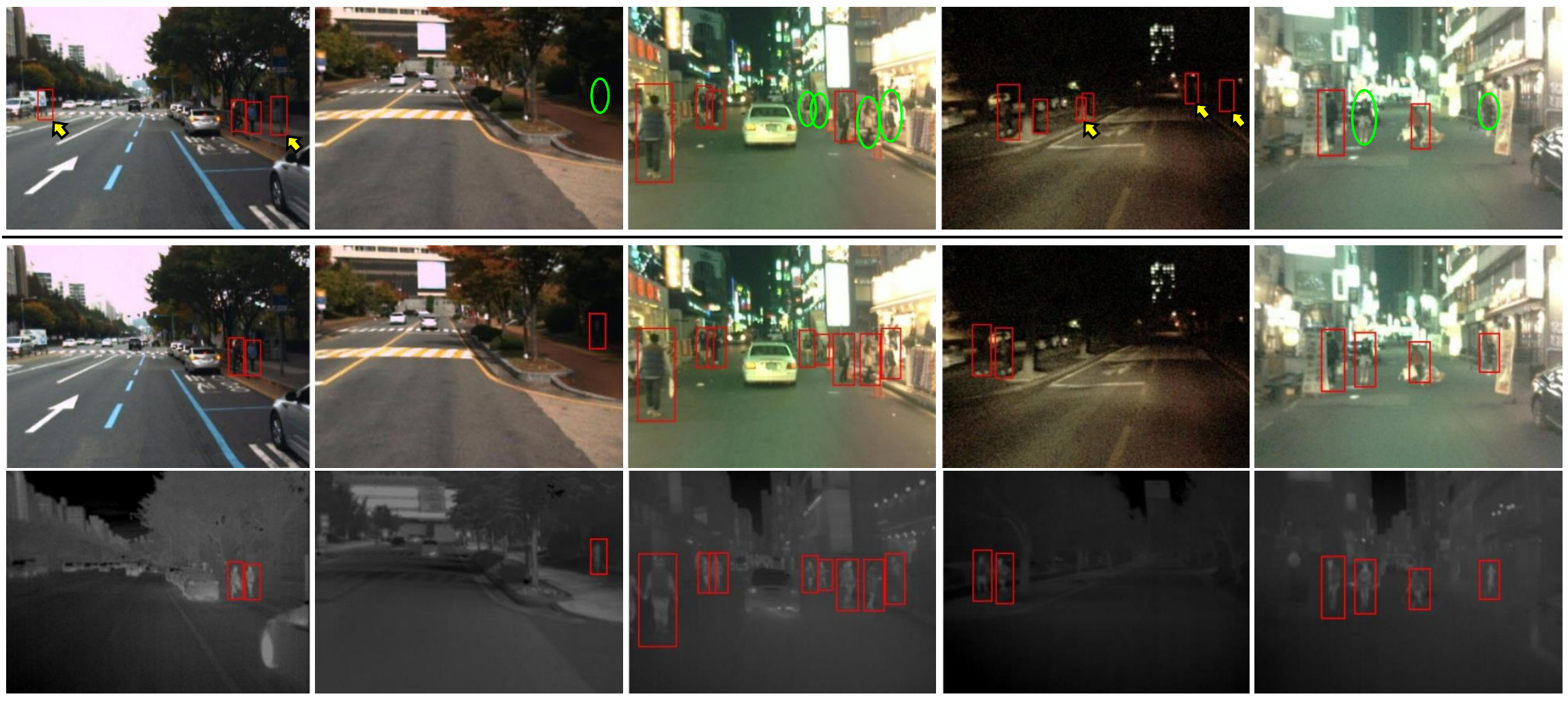}
\vspace{-5pt}
\caption{Detection samples. Red bounding boxes denote detections. Yellow arrows indicate false positives and green ellipses represent miss detections. First row: detections by FasterRCNN-C (color images only). Bottom two rows: detections by Halfway Fusion model (multispectral images), illustrated in both color and thermal images.}
\label{fig:results}
\end{figure}

Among the four ConvNet fusion models, Halfway Fusion achieves the lowest overall MR (36.99\%), which is 3.5\% lower than other fusion models, showing the most effective multispectral synergy for pedestrian detection. Since the four DNNs architectures correspond to information fusion on different ConvNet levels, we speculate that middle-level convolutional features from color and thermal branches are more compatible in fusion: they contain some semantic meanings and meanwhile do not completely throw all fine visual details. However, Early Fusion combines low-level features, such that some task irrelevant low-level features would be fused, which could undermine the fusion power. Late Fusion executes fusions on high-level semantic features and Score Fusion combines confidence scores. In some cases, it would be difficult for these two models to eliminate semantic noise or to adjust decision mistake of one image channel. Compared to FasterRCNN-C or FasterRCNN-T, given the whole test set, Halfway Fusion reduces the overall MR by around 11\%,  

Some detection samples are illustrated in Figure~\ref{fig:results}. Detections with confidences large than 0.5 are presented. Obviously, compared to the color image based detector, our multispectral pedestrian detector achieves more true detections, especially when some pedestrians are of bad external illumination. Meanwhile, some false alarms are also removed.

\subsection{Evaluation of Proposals:}
\begin{figure}[t]
\centering
\subfigure{
\includegraphics[width=0.33\textwidth]{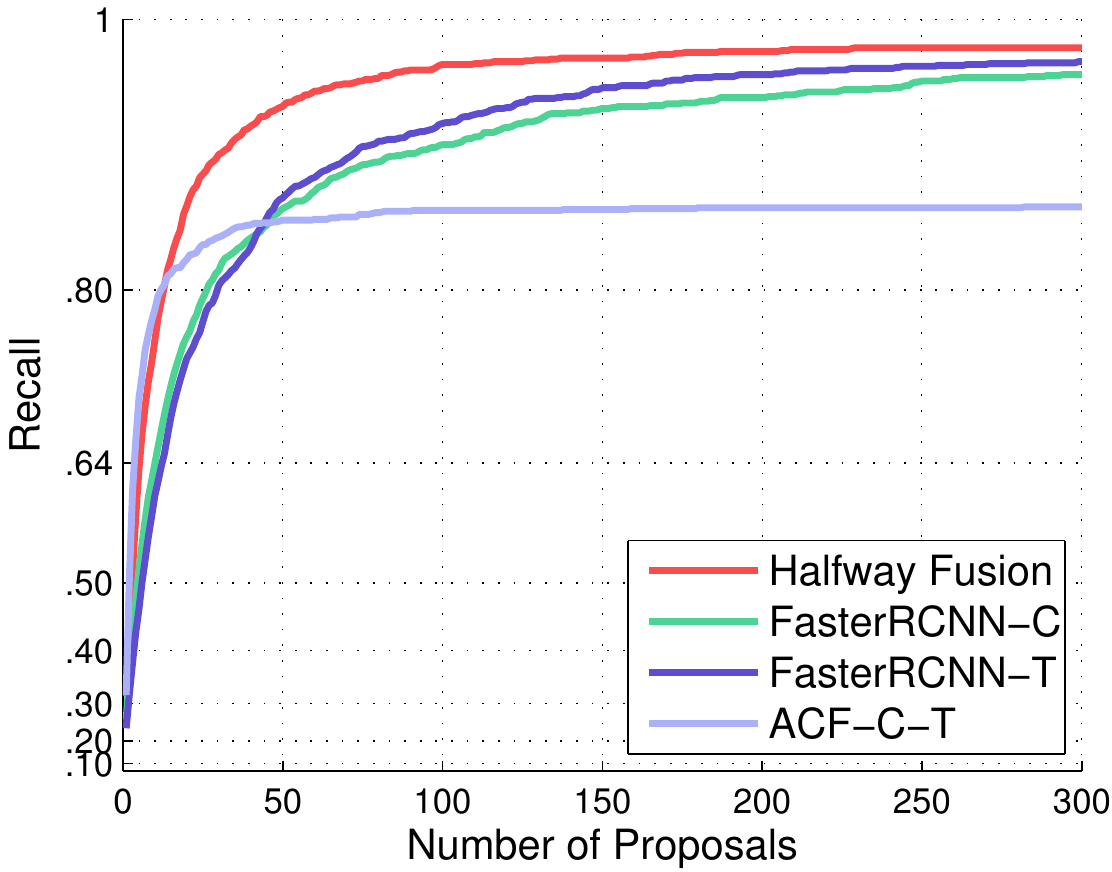}}
\hspace{5pt}
\subfigure{
\includegraphics[width=0.33\textwidth]{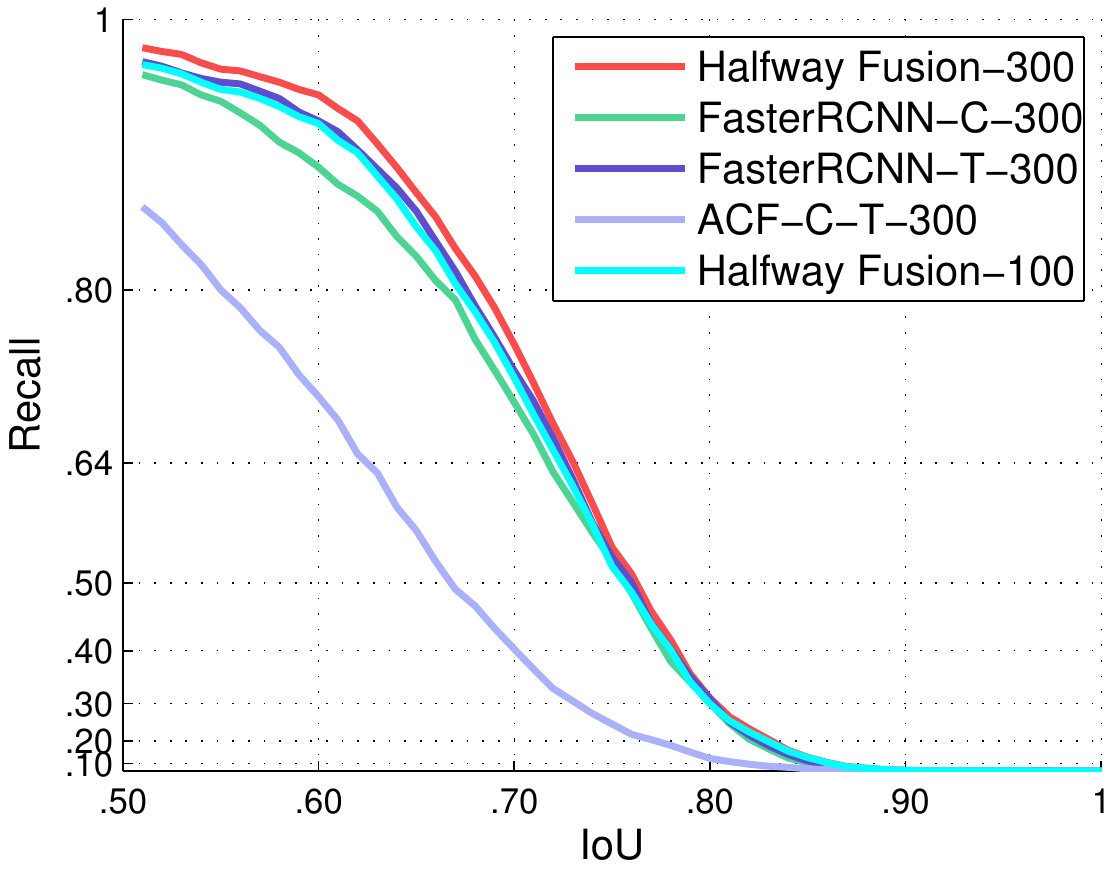}}
\caption{Comparison of pedestrian proposals reported on test set of KAIST multispectral pedestrian dataset. Left: Recall \vs Number of Proposals; Right: Recall \vs IoU.} 
\label{fig:recall}
\vspace{-6pt}
\end{figure}

We also assessed the proposals generated by RPN in the Halfway Fusion, with regard of recalls. RPNs of FasterRCNN-C and FasterRCNN-T were considered in comparison, along with the ACF-C-T pedestrian detector. The comparison results of pedestrian proposals performed on the test set of KAIST are shown in Figure~\ref{fig:recall}.

\myparagraph{Recall \vs Number of Proposals:} Given IoU 0.5, Halfway Fusion model obtains the highest recall with the same number of proposals. This model achieves 94\% recall with 50 proposals, compared to other approaches of around 87\% recall. In other words, Halfway Fusion could reach the same recall with fewer proposals. This is very useful in practice, since fewer proposals could make DNNs save time in classification. In particular, Halfway Fusion gets 90\% recall with 30 proposals, while FasterRCNN-C and FasterRCNN-T require around 80 proposals to achieve competitive recall. 

\myparagraph{Recall \vs IoU:} Given 300 proposals from RPN, Halfway Fusion obtains 93.9\% recall at IoU 0.6, which is better than other approaches. With 100 proposals, Halfway Fusion model accomplishes comparative recalls against other methods with 300 proposals. Apparently, Halfway Fusion model produces proposals with better overlaps on true detections. In this scenario, we conclude that Halfway Fusion model generates proposals of higher quality.

\section{Conclusion}
\label{sec:conclude}
In this paper, we focused on leveraging DNNs for multispectral (color and thermal images) pedestrian detection. Our multispectral detectors were built upon Faster R-CNN detection framework, which archived the state-of-art performance on Caltech pedestrian benchmark. Four ConvNet fusion architectures were proposed, which fused channel features at different ConvNet stages, corresponding to low-level, middle-level,  high-level feature fusion, and confidence fusion, respectively.
All of them yielded better performance compared with the baseline detector based on Faster R-CNN. 
Extensive empirical results revealed that our Halfway Fusion model -- the fusion of middle-level convolutional features, achieved the best detection synergy and the state-of-the-art performance.
It significantly reduced the missing rate of baseline method Faster R-CNN by 11\%, and performed a 37\% overall missing rate (3.5\% lower than other proposed architectures) on KAIST datasets.

\bibliography{egbib}
\end{document}